\pdfoutput=1

\documentclass[11pt]{article}

\usepackage{ACL2023}

\usepackage{times}
\usepackage{latexsym}
\usepackage{comment}
\usepackage{booktabs}
\usepackage[T1]{fontenc}

\usepackage[utf8]{inputenc}

\usepackage{microtype}

\usepackage{inconsolata}

\usepackage{graphicx}

\title{Team UTSA-NLP at SemEval 2024 Task 5: Prompt Ensembling for Argument Reasoning in Civil Procedures with GPT4}

\author{Dan Schumacher$^1$ \and Anthony Rios$^2$ \\
  $^1$MS Data Analytics, College of Business \\
  $^2$Department of Information Systems and Cyber Security\\
  The University of Texas at San Antonio \\
  \texttt{dan.schumacher@my.utsa.edu, anthony.rios@utsa.edu} }

\begin{document}
\maketitle
\begin{abstract}

In this paper, we present our system for the SemEval Task 5, \textit{The Legal Argument Reasoning Task in Civil Procedure} Challenge. Legal argument reasoning is an essential skill that all law students must master. Moreover, it is important to develop natural language processing solutions that can reason about a question given terse domain-specific contextual information. Our system explores a prompt-based solution using GPT4 to reason over legal arguments. We also evaluate an ensemble of prompting strategies, including chain-of-thought reasoning and in-context learning. Overall, our system results in a Macro F1 of .8095 on the validation dataset and .7315 (5th out of 21 teams) on the final test set. Code for this project is available at \url{https://github.com/danschumac1/CivilPromptReasoningGPT4}.

\end{abstract}

\section{Introduction}

Mastering the reasoning behind legal arguments is a fundamental skill required of all law students. In this study, we develop a novel approach for SemEval Task 5, The Legal Argument Reasoning Task in Civil Procedure Challenge~\cite{held-habernal-2024-legalreasoning}. The SemEval Task released a dataset that was scraped from \textit{The Glannon Guide To Civil Procedure}, a textbook designed for law students. Specifically, given case law, a question, and a potential answer to that question, students must be able to reason over the contextual information (case law) to determine if the question is correct or not. 


There has been substantial research in developing NLP-based reasoning systems~\cite{guha2024legalbench,bongard2022legal,chalkidis2023chatgpt,blair2023can,kuppa2023chain,yu2022legal}. The methods can be categorized into two major frameworks: fine-tuning and large language model-based (LLM) solutions. For fine-tuning approaches, \citet{bongard2022legal} introduced an approach that fine-tunes LegalBERT~\cite{chalkidis2020legal} and developed several methods for handling long text that does not fit within the token limitations of LegalBERT. For LLM solutions, \citet{chalkidis2023chatgpt} explored the use of ChatGPT for solving legal exams. \citet{guha2024legalbench} introduced a more comprehensive evaluation benchmark geared to large language models consisting of 162 tasks. Many of their experiments show that GPT4 is one of the top performing approaches for legal reasoning across all language models, while Flan-T5-XXL is the best open-source option. Although showing substantial generalization is important, developing specific prompting strategies for different reasoning tasks can substantially improve performance. Hence, this paper adds to existing literature on the exploration of prompting approaches in the legal domain.

For this work, we adapt several prompting-based strategies to develop an LLM-based solution for the shared task. Specifically, we combine a retrieval system with in-context learning and chain-of-thought reasoning. There are several studies showcasing the utility of in-context learning~\cite{liu2022makes,lu2022dynamic} and chain-of-thought reasoning~\cite{wei2022chain,sun2023text,yao2024tree}. Moreover, there is work using both human-curated and machine-generated reasons. In this work, we focus on human-generated reasons for the training data, and machine-generated examples are only used at test times when human expert annotations are not provided. Furthermore, rather than providing a step-by-step reasoning approach, which may not make sense in this context, our approach is more similar to single-step rationales~\cite{brinner2023model,yasunaga2023large}, which simply provides a single step of reasoning for why an answer is correct or incorrect.

In summary, this paper makes the following contributions to our solution for the SemEval 2024 Task 5 shared task:
\begin{itemize}
    \item We evaluate prompting strategies using GPT4 that combine several popular ideas, including in-context learning and chain-of-thought reasoning.
    \item In-context learning can be sensitive to the actual choice of examples, particularly when only a few examples are provided. Hence, we also explore an ensemble of prompt-based predictions to improve overall performance.
    \item Finally, we provide a unique error analysis where we found limitations and common error types generated by GPT4 using our prompting strategies. For example, when a part of an answer candidate is correct, but the reasoning is wrong, GPT4 is likely to generate a false positive (i.e., predict it is correct instead of incorrect).
\end{itemize}


\section{RELATED WORK}

Overall, there are three major areas of legal NLP: legal question-answering~\cite{khazaeli2021free,kien2020answering,ryu2023retrieval,martinez2023survey,wang2023maud}, judgment prediction~\cite{masala2021jurbert,valvoda2023role,juan2023custodiai}, and corpus mining (e.g., summarization, text classification, information extraction, and retrieval-related research)~\cite{poudyal2020echr,li2022detecting,vihikan2021automatic,zhang2023diverse,limsopatham2021effectively,de2023bb25hlegalsum}. There has also been some broad methodology work that is aimed at working on various legal tasks in general (e.g., LegalBERT~\cite{chalkidis2020legal}). 

The SemEval task is most similar to question-answering related research. In the domain of legal question-answering, recent research efforts are focused on creating new systems, developing evaluation criteria, and compiling datasets, considering the significant variation across different legal fields. \citet{khazaeli2021free} introduced a commercial question-answering system for legal inquiries, leveraging information retrieval techniques, sparse vector search, embeddings, and a BERT-based re-ranking system, trained on both general and legal domain data. \citet{ryu2023retrieval} developed a novel evaluation method for LLM-generated texts that assess their validity using retrieval-augmented generation, showing improved alignment with legal experts' assessments and effectiveness in identifying factual errors. \citet{wang2023maud} created the Merger Agreement Understanding Dataset (MAUD), a unique, expert-annotated dataset for legal text reading comprehension, highlighting promising model performance and the need for further improvement in understanding complex legal documents.

From a methodological point-of-view instead of a task-oriented view, recent research efforts have concentrated on the advancement of NLP-based reasoning systems, with a particular focus on applications within the legal domain~\cite{guha2024legalbench,bongard2022legal,chalkidis2023chatgpt,blair2023can,kuppa2023chain,yu2022legal}. These efforts can be broadly classified into two distinct methodologies: fine-tuning approaches and those leveraging large language models (LLMs). Within the fine-tuning paradigm, \citet{bongard2022legal} have proposed modifications to LegalBERT~\cite{chalkidis2020legal} aimed at enhancing its ability to process texts that exceed the model's inherent token limitations. This approach is representative of a broader trend towards tailoring pre-existing models to better suit specific textual analysis tasks in the legal sector. \citet{kien2020answering} developed a retrieval-based model employing neural attentive text representation with convolutional neural networks and attention mechanisms for accurately matching legal questions to relevant articles, demonstrating superior performance on a Vietnamese legal question dataset.

Conversely, the exploration of LLMs has also been wide covering new general approaches for legal question answering and reasoning to new datasets and benchmarks.  \citet{yu2023exploring}  investigated the impact of chain-of-thought prompts and fine-tuning methods on legal reasoning tasks, specifically the COLIEE entailment task, and found that prompts based on legal reasoning techniques and few-shot learning with clustered training data significantly enhance performance. \citet{chalkidis2023chatgpt} demonstrates the potential of utilizing models like ChatGPT for complex reasoning tasks, such as solving legal examination questions. Building on this, \citet{guha2024legalbench} have introduced a comprehensive evaluation framework designed specifically for assessing the capabilities of LLMs across a suite of 162 tasks. This benchmark aims to provide a more nuanced understanding of the strengths and limitations of LLMs in the context of legal reasoning. Additionally, while employed within a distinct domain from legal reasoning, a comparable methodology in prompt engineering—encompassing chain of thought prompting and in-context prompting—is illustrated in Liu et al.'s recent work \cite{liu2022interpretable}. Overall, compared to the prior work that developed methods and datasets for generating answers to questions, the SemEval task focuses on understanding whether a provided answer candidate is valid given a specific context.

\section{METHOD}

\begin{figure*}
    \centering
    \includegraphics[width=.85\linewidth]{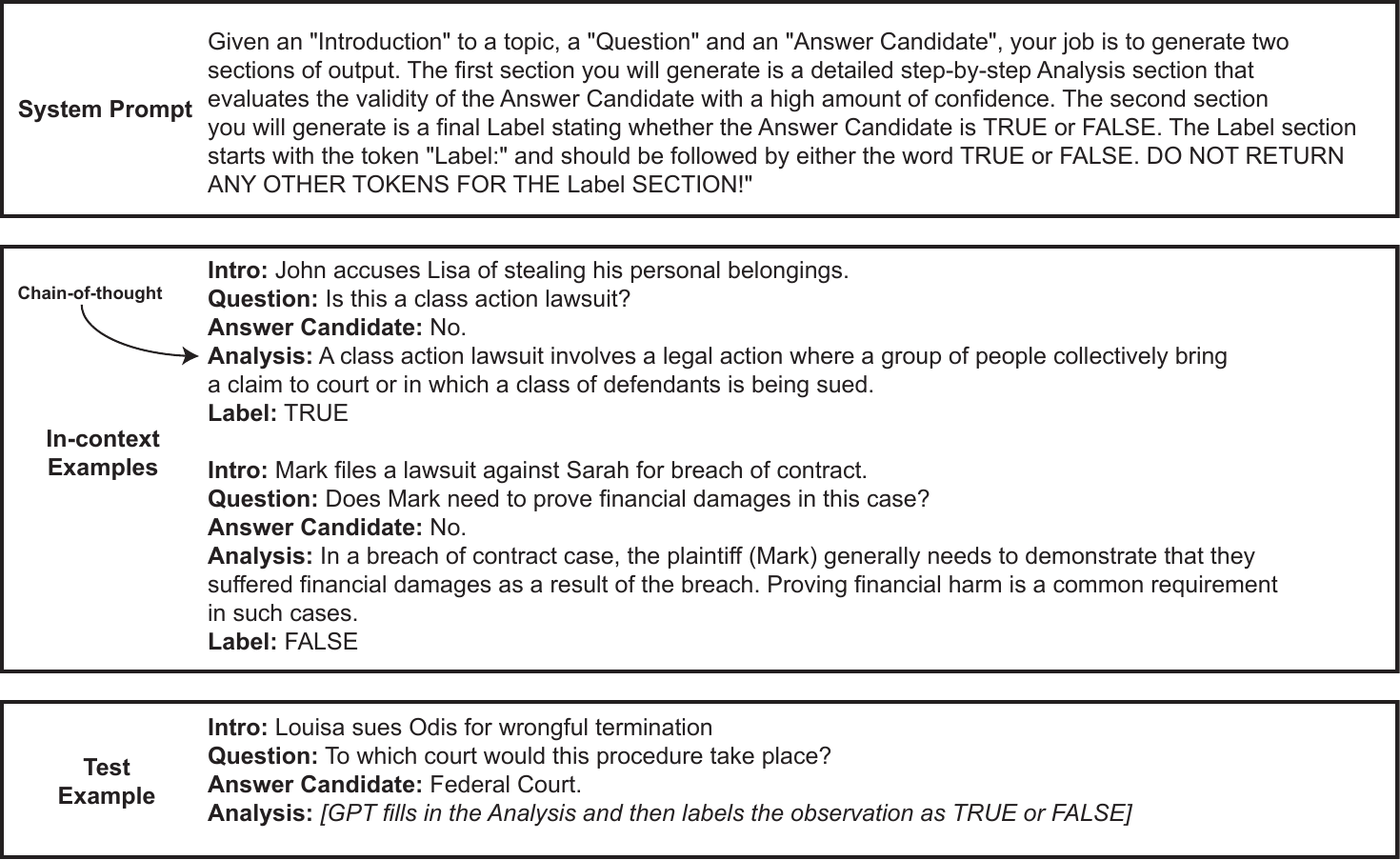}
    \caption{Overview of our prompting strategy. }
    \label{fig:overview}
    \vspace{-1em}
\end{figure*}

We provide a high-level overview of our approach in Figure~\ref{fig:overview}. Overall, we explore three major different prompting approaches: Zero-shot prompting, few-shot prompting, and few-shot prompting with chain-of-thought-like reasoning. Moreover, we explore an ensemble of multiple approaches. The methods are described in the following subsections.

\subsection{Few-Shot Retrieval Augmented Chain-of-Thought Prompting}

We refer to our approach as ``Few-Shot \& CoT \& RAG.'' Each example within the dataset contains an Introduction, Question, Answer Candidate, Analysis, and Label. For new examples at test time, we only have access to the Introduction, Question, and Answer Candidate. The Introduction consists of a general background about a legal case. The Question is about the case, and the Answer Candidate is an answer to the Question. It is important to note that the Question could be in question form, where the answer directly answers what is asked. However, it may function as a fill-in-the-blank exercise, where the question presents an incomplete statement that the Answer Candidate is expected to complete. The Analysis, which is only provided in the training and development datasets, is a detailed expert-defined explanation for why the Answer Candidate is or is not valid. The Label is a TRUE or FALSE value, where TRUE means that the Answer Candidate correctly addresses the question given the provided context. Likewise, FALSE means that the Answer Candidate is incorrect.

As shown in Figure~\ref{fig:overview}, our prompting strategy contains three main components, a system prompt, the in-context examples, and the final test instance we will classify as TRUE or FALSE. The system prompt describes what the large language model (LLM) should do. In the Figure, it is shown that we also added explicit information to limit the model from generating non-relevant information.

The in-context examples are provided to the LLM before the final text instance. Intuitively, the goal is to provide some examples of the task we are accomplishing to better ground the LLM to make better-generated responses. This is the Retrieval Augmented aspect of our system (i.e., RAG). While any random examples could be provided, we search for the most relevant examples for each test instance. Formally, given an input instance $x_i$ consisting of a concatenated Introduction, Question, and Answer triplet $w$, we retrieve the most similar examples $\{x_1, \ldots, x_{\mathcal{N}(x_i)}\}$, where $\mathcal{N}(x_i)$ are the $k$ most similar examples to $x_i$. Each question-answer pair is embedded using LegalBERT.
The in-context examples all come from the provided training dataset. Once retrieved, all of the relevant information for the retrieved examples are used in the prompt (i.e., the Introduction, Question, Answer Candidate, Analysis, and Label). Figure~\ref{fig:overview} shows an example with two in-context examples.

Finally, the last component of the prompt is the text example. Basically, for every example we wish to make a prediction for, we pass the Introduction, Question, and Answer Candidate. The model will first generate the Analysis for that example, then it will generate the Label.

\subsection{Ensemble}

Besides the method described in the previous section, we also explored prompting variants to create an ensemble. We describe each of the additional methods below (besides the Few-SHOT \& CoT \& RAG  method described in the previous subsection). 

\paragraph{Zero-shot.} This approach does not use any in-context examples. We provide the system prompt and the test example to the model directly to get the final prediction.

\paragraph{Zero-shot \& CoT.} This method builds on the Zero-Shot approach by adding the CoT aspect to the test example. Note that this is not available at test time, so the model generates an Analysis section without previous examples.

\paragraph{Few-Shot.} The few-shot approach will use multiple in-context examples. However, unlike our main method explained in the previous section, it uses the same in-context examples for all test cases. The examples were chosen in an ad-hoc manner with an emphasis on relatively short examples to limit the number of tokens to reduce costs.

\paragraph{Few-Shot \& CoT.} This builds on the Few-Shot method, where ad-hoc in-context examples are still used, but we also provide the CoT reasoning (Analysis) section to the in-context examples. The GPT4 model will generate the Analysis section for each test example.

\paragraph{Few-Shot \& RAG.} This also builds on the Few-Shot method, but instead of using ad-hoc examples, it uses LegalBERT and cosine similarity to find relevant in-context examples for each test case (as explained in the previous section).

Overall, there are a total of 4 models in our ensemble. Few-Shot and Few-Shot \& RAG were not used, but we evaluated them. The models in the ensemble were chosen by checking combinations on the validation dataset. To make a prediction, we use voting with a threshold (i.e., where the votes are processed to generate the proportion of TRUE values).


\subsection{Model Details}

 For all of our prompts, we use GPT-4-1106-preview with a temperature of .7. The similarity metric used for finding relevant in-context examples is cosine similarity. Moreover, we choose a total of 2 in-context examples, consisting of 1 TRUE example and 1 FALSE example. We searched for the best threshold after voting by calculating the proportion of TRUE vs. FALSE predictions, ultimately choosing .5 as the threshold.

There were many cases where GPT4 does not return a final label in an easy-to-process fashion (i.e., it does not end with a TRUE or FALSE). We explored multiple approaches to parse the answer.  Our ultimate strategy involves re-submitting examples to GPT-4 that initially failed to produce a valid label, explicitly indicating the absence of the label, and prompting the model to generate the correct information. 
\section{RESULTS}

\begin{table}[t]
\centering
\resizebox{\columnwidth}{!}{%
\begin{tabular}{lrr}
\toprule
\textbf{Method}                                                          & \textbf{Macro F1} & \textbf{Acc.} \\ \midrule
\multicolumn{3}{c}{Baselines} \\ \midrule
RoBERTa & .5128	& .2286 \\
Legal-BERT & .5575	 & .2941 \\ \midrule
Zero-shot                                                                         & .6681            & .7857            \\
Zero-Shot \& COT                                                                  & .7162            & .7500            \\
Few-shot                                                                         & .6935            & .7738            \\
Few-shot \& COT                                                                  & .6762            & .7262            \\
Few-shot \& RAG                                                                  & .6898            & .7500            \\ \midrule
Few-Shot \& COT \& RAG  (Ours)                                                        & .7306            & .7857            \\ 
Ensemble (Ours) & .8095            & .8571            \\ \bottomrule
\end{tabular}%
}
\caption{Validation dataset results}
\vspace{-2em}
\label{tab:val}
\end{table}

In this section, we present our own results on the validation dataset as well as the final results in the competition.

\noindent \textbf{Baselines}
In our experiments, using the validation data, we compare our approach (Few-SHOT \& CoT \& RAG) with the other approaches used in the Ensemble. Moreover, we compare our system to both RoBERTa~\cite{liu2019roberta} and Legal-BERT. However, because both RoBERTa and Legal-BERT are limited to 512 tokens, we split the Introduction into $b$ pieces. $b$ is calculated by subtracting the number of words in the question and answer from 512. We halve that result to accommodate the fact that the number of tokens typically exceeds the number of words. Next, we divide the number of words in the Introduction by the previously calculated number to obtain the total number of windows. Finally, all of the tokens in the Introduction are evenly split into the windows. Each piece of the introduction is appended to the Question and Answer pair independently to generate multiple predictions for each instance. We then use voting to make a final prediction for the entire sequence. This method is similar to what was explored in \citet{bongard2022legal}.

\noindent \textbf{Validation Results.}
The validation performances are shown in Table~\ref{tab:val}. We observe that the fine-tuned methods (e.g., RoBERTa), perform less effectively compared to all methods utilizing GPT-4. Between the fine-tuned methods, we find that Legal-BERT outperforms RoBERTa, which is expected given Legal-BERT was fine-tuned on relevant corpora.

Next, between GPT4 methods (not including the ensemble), we find that performance varies substantially between .6681 and .7162 for Macro F1. All methods outperform Zero-Shot. However, Zero-Shot \& COT achieved the best performance across all baseline methods for Macro-F1. When we compare the baseline approaches to our method (Few-Shot \& COT \& RAG), we find that the method outperforms all variations. From an ablation standpoint, removing RAG has the biggest performance drop (.7306 vs. .6762). Interestingly, we find that removing CoT has the second largest drop in performance (.7306 vs. .6935), and removing the in-context examples has the smallest drop in performance (.7306 vs. .7162). Before the study, we expected removing the in-context examples would result in the largest performance drop.

\begin{table}[t]
\centering
\resizebox{.7\columnwidth}{!}{%
\begin{tabular}{@{}llll@{}}
\toprule
\textbf{\#} & \textbf{User}            & \textbf{Macro F1} & \textbf{Acc.} \\ \midrule
1  & zhaoxf4         & .8231   & .8673   \\
2  & irene.benedetto & .7747   & .8265   \\
3  & kbkrumov        & .7728   & .8367   \\
4  & qiaoxiaosong    & .7644   & .8163   \\
\textbf{5}  & \textbf{UTSA-NLP}        & \textbf{.7315 }  & \textbf{.7959}   \\
6  & kubapok         & .6971   & .7857   \\
7  & samyak          & .6599   & .7449   \\
8  & hrandria        & .6327   & .6939   \\
9  & Yuan\_Lu        & .6000   & .6327   \\
10 & PengShi         & .5910   & .6735   \\
11 & msiino          & .5597   & .5714   \\
12 & Hwan\_Chang     & .5556   & .5918   \\
13 & kriti7          & .5511   & .6020   \\
14 & woody           & .5510   & .6633   \\
15 & odysseas\_aueb  & .5143   & .6122   \\
16 & Manvith\_Prabhu & .4966   & .6224   \\
17 & lhoorie         & .4957   & .5000   \\
18 & yms             & .4827   & .7245   \\
19 & U\_201060       & .4503   & .6633   \\
20 & langml          & .4375   & .4490   \\
21 & lena.held       & .4269   & .7449   \\ \bottomrule
\end{tabular}%
}
\caption{Final Competition Results. Our submission is in \textbf{bold} font.}
\vspace{-1em}
\label{tab:comp}
\end{table}

\noindent \textbf{Competition Results.}
In Table~\ref{tab:comp}, we report the final results of the competition, achieving a Macro F1 of .7315. But, why do we see such a large performance drop between the competition and validation results (.7315 vs. .8095)? We hypothesize two major reasons. First, we realized that the validation dataset contains many Introduction-Question pairs identical to the ones in the training dataset. Despite the Answer Candidates differing across the two datasets, the substantial overlap in Introduction-Question pairs may lead to an overestimation of our model's performance on the validation dataset, rendering it potentially too optimistic when applied to entirely new data. Second, we spent time over-optimizing the ensemble on the validation dataset causing overfitting issues (e.g., checking thresholds, model combinations, and more). By generating a better validation split, we may have seen better generalization.

\begin{table}[t]
\centering
\resizebox{.7\columnwidth}{!}{%
\begin{tabular}{@{}lll@{}}
\toprule
              & \textbf{predFalse} & \textbf{predTrue} \\ \midrule
\textbf{actFalse} & 57                  & 9\\
\textbf{actTrue}  & 3& 15                \\ \bottomrule
\end{tabular}}
\caption{Confusion matrix for the validation dataset. actFalse and actTrue stand for actual True and False values, respectively. predFalse and predTrue stand for predicted False and predicted True.}
\vspace{-2em}
\label{tab:confusion}
\end{table}

\noindent \textbf{Error Analysis}
Our method resulted in twelve mistakes on the validation dataset: ten false positives and two false negatives. The confusion matrix is shown in Figure~\ref{tab:confusion}. In Table~\ref{tab:statisticserror} (See Appendix), we categorize each false positive into one of 
 four categories: ``Incorrect reasoning,'' ``Shared the same introduction and question pair,'' ``lots of similar language'', and ``Other.'' With only two false negatives, there are few useful patterns to understand among the errors. Hence, we only have a general FN category. However, from the false positives, we make two major findings which we describe below.

The first was when the answer candidate had the correct answer but \textit{incorrect reasoning}. Here is a toy example demonstrating this pattern:
\begin{quote}
\textbf{Introduction}: Carlos enjoys riding his skateboard in the skate park. Unfortunately, during one of his rides, he fell and split his head open.

\textbf{Question}: Should Carlos go to the hospital?

\textbf{Answer Candidate}: Yes, Carlos should go to the hospital because he likes to kick-flip so much.

\textbf{Analysis}: While it is correct that Carlos should go to the hospital for medical attention after splitting his head open, the reasoning provided in the answer candidate is flawed. The decision to seek medical help should be based on the severity of the injury and the need for professional medical treatment, not on Carlos's enjoyment of skateboard stunts.
 \end{quote}
Intuitively, part of the Answer Candidate is correct, i.e., Carlos \textit{should} go to the hospital, yet the reasoning that states why he should go to the hospital is wrong.

The second point is that three of our ten false positives all \textit{shared the same introduction and question pair.} The introduction contained more extraneous information than usual and was 128  words longer than the average. In these instances, our model would analyze the answer candidate as correct but without taking into account the particular case that was asked about. Below is a toy example:
\begin{quote}
\textbf{Introduction}: My dog Louisa loves to learn new tricks, go for walks, eat her dinner, then sleep through the night

\textbf{Question}: What does Louisa like to do \textit{after dinner}?

\textbf{Answer Candidates}:
\begin{enumerate}
    \item  Learn new tricks (\textit{wrong})
    \item  Go for walks (\textit{wrong})
    \item  Eat her dinner (\textit{wrong})
    \item Sleep through the night (\textit{correct})
\end{enumerate}
\end{quote}
In the example, three of the answers are incorrect, while ``Sleep through the night'' is the correct example. When the Introduction is long, the actual context of the question may be ignored, which can be interpreted as a needle in a haystack issue. Developing better systems that provide direct ``attention'' to relevant information may improve performance.

Finally, a third point that we believe caused our model to predict incorrect answer candidates is \textit{similar language} in both the introduction and question compared to the answer candidate. 
\begin{quote}
\textbf{Introduction:}
While making eggplant Parmesan for the first time in a buttery dutch over Paige burned herself on the stove

\textbf{Question: }
How does Paige remember this incident.

\textbf{Answer Candidates:}
She remembers making eggplant Parmesan for the first time in a buttery dutch oven fondly.
\end{quote}

\section{CONCLUSION}

In this paper, we described our approach for 2024 SemEval Task 4, The Legal Argument Reasoning Task in Civil Procedures. Specifically, we introduced a GPT4 prompting-based strategy that achieved 5th place in the competition out of 21 participants. Overall, we find that combining in-context learning, where we use a retrieval-based approach to find relevant examples, as well as in-context learning improves model performance. Based on our experiments, there are three natural areas for future research. First, the actual Analysis section used for chain-of-thought reasoning does not match traditional methods which use step-by-step reasoning. Hence, a logical next extension is to reword (potentially with GPT4) the Analysis section to provide a step-by-step explanation for an answer. Second, this work was limited to 2 in-context examples to limit API costs and allow us to test other models in our initial experiments. However, extending that to 10 or more examples can potentially improve performance. Third, the current approach relies on a closed-source model (GPT4). Exploring open-source models, particularly smaller open-source models such as T5~\cite{chung2022scaling} and LLama2~\cite{touvron2023llama}, is important to better understand the impact of pretraining data on performance.

\section*{ACKNOWLEDGEMENTS}

This material is based upon work supported by the National Science Foundation (NSF) under Grant~No. 2145357.

\bibliography{anthology,custom}
\bibliographystyle{acl_natbib}

\appendix

\begin{table*}[t]
\centering
\renewcommand{\arraystretch}{1.2}
\resizebox{\linewidth}{!}{%
\begin{tabular}{llp{9cm}}

\toprule
              \textbf{Type}& \textbf{Frequency}& \textbf{Example}\\ \midrule
FP: Incorrect Reasoning& 16.67\% (1/12)& Carlos enjoys riding his skateboard in the skate park...\\
 FP: Shared the same introduction and question pair.& 25\% (3/12)&My dog Louisa loves to learn new tricks...\\
 FP: Lots of Similar Language& 8.33\% (2/12)& While making Eggplant Parmesean...\\
 FP: Other& 25\% (3/12)&...\\
FN& 25\% (3/12)& ...\\ \bottomrule
\end{tabular}}
\caption{Manual categorization of error types. False positives are categorized as either ``Incorrect Reasoning,'' ``Shared the same introduction and question pair,'' ``lots of similar language,'' or ``Other.'' We use a single FN category because of the lack of errors to analyze.}
\label{tab:statisticserror}
\end{table*}

\section{Error Analysis}

Table~\ref{tab:statisticserror} reports the basic statistics for the error types we observed in our error analysis.
\end{document}